\pdfoutput=1

\documentclass[11pt]{article}

\usepackage{emnlp2021}

\usepackage{times}
\usepackage{latexsym}

\usepackage[T1]{fontenc}

\usepackage[utf8]{inputenc}

\usepackage{microtype}

\usepackage{graphicx}
\usepackage{caption,subcaption}

\usepackage{amsmath}
\usepackage{booktabs}
\usepackage{multirow}
\usepackage{ulem}

\usepackage{xcolor}

\usepackage{soul}

\newcommand{\myul}[2][black]{\setulcolor{#1}\ul{#2}\setulcolor{black}}

\newcommand{\comment}[3]{{\small{\textcolor{#3}{[#1 #2]}}}}

\newcommand{\com}[1]{}

\newcommand{\marker}[1]{#1:}

\newcommand{\roy}[1]{\comment{\marker{ROY}}{#1}{orange}}

\newcommand{\yonatan}[1]{\comment{\marker{YONATAN}}{#1}{blue}}

\newcommand{\resolved}[1]{}

\newcommand{\lossgap}[0]{\textit{$\Delta$ image loss}}
\newcommand{\originalstrategy}[0]{Baseline MLM}
\newcommand{\cwstrategy}[0]{Content words}
\newcommand{\repourl}[0]{https://github.com/yonatanbitton/data_efficient_masked_language_modeling_for_vision_and_language}

\title{Data Efficient Masked Language Modeling for Vision and Language}

\newcommand{\authorspace}[0]{\quad}

\author{
  Yonatan Bitton$^{\diamondsuit}$ \authorspace 
 Gabriel Stanovsky$^{\diamondsuit}$
 \authorspace
 Michael Elhadad$^{\spadesuit}$\authorspace 
 Roy Schwartz$^{\diamondsuit}$\\
 $^{\diamondsuit}$School of Computer Science and Engineering, The Hebrew University of Jerusalem, Israel \\
 $^{\spadesuit}$ Department of Computer Science, Ben Gurion University, 
   Israel \\
  \{yonatanbitton,gabis,roys\}@cs.huji.ac.il \authorspace elhadad@cs.bgu.ac.il 
}

\begin{document}
\maketitle

\begin{abstract}
Masked language modeling (MLM) is one of the key sub-tasks in vision-language pretraining.
In the cross-modal setting, tokens in the sentence are masked at random, and the model predicts the masked tokens given the image and the text. In this paper, we observe several key disadvantages of MLM in this setting. First, as captions tend to be short, in a third of the sentences no token is sampled. Second, the majority of masked tokens are stop-words and punctuation, leading to under-utilization of the image.
We investigate a range of alternative masking strategies specific to the cross-modal setting that address these shortcomings, aiming for better fusion of text and image in the learned representation.
When pre-training the LXMERT model, our alternative masking strategies consistently improve over the original masking strategy on three downstream tasks, especially in low resource settings. Further, our pre-training approach substantially outperforms the baseline model on a prompt-based probing task designed to elicit image objects. These results and our analysis indicate that our method allows for better utilization of the training data.\footnote{Our code, pre-trained, and fine-tuned models are published at \url{\repourl}.}
\end{abstract}
\normalem
\section{Introduction}
\begin{figure}[!ht]
\centering
\newcommand{\figlen}[0]{\columnwidth}
    \includegraphics[width=\figlen]{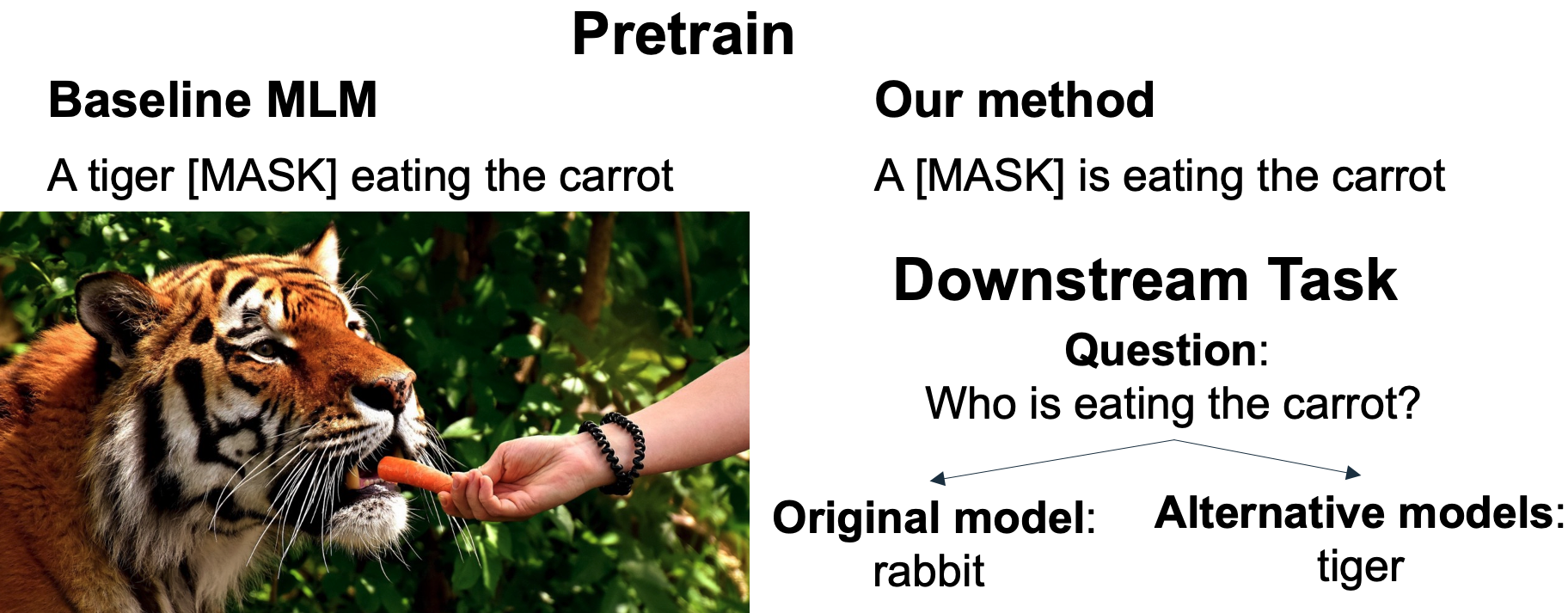}\\
\caption{Illustration of our approach. The baseline MLM masks a random token with 15\% probability, where $\approx$50\% of the masked tokens are stop-words or punctuation. Our method masks words that require the image in order to be predicted (e.g., physical objects).
Our pre-train masking strategy consistently improves over the baseline strategy in two evaluation setups. 
}
    \label{fig:fig1}
\end{figure}

Pre-trained vision-language (VLP) models such as ViLBERT \cite{lu2019vilbert}, LXMERT \cite{tan-bansal-2019-lxmert} and UNITER \cite{chen2020uniter} have recently improved the state-of-the-art across various vision and language benchmarks.
One of the primary pre-training objectives of VLP is masked language modeling (MLM). Motivated by the single-modal MLM task, most models perform as introduced in BERT \cite{devlin2018bert} for text-only data, randomly masking tokens with a probability of 15\% \cite{shin2021perspectives}. 

The main difference in the cross-modal setting\footnote{This task is often referred to as ``cross-modality MLM'', or ``MLM conditioned on image regions'' \cite{chen2020uniter}, to emphasize the difference from the text-only MLM task.} is that the model takes into account both the textual context and the image, and the latter can help it resolve ambiguities. For example, in Figure~\ref{fig:fig1}, given the masked sentence “A [MASK] is eating the carrot”, without the image, the model might predict \textit{rabbit}, since it is correlated with \textit{carrot}. But the image reveals that the answer is \textit{tiger}.

In this work, we find that the  MLM pre-training method is sub-optimal for VLP, as it does not make efficient use of the training data. This manifests in two major shortcomings, common to many popular pre-train datasets \cite{lin2014microsoft, krishna2017visual, sharma2018conceptual, ordonez2011im2text}. First, we observe that image captions, which form the textual part of these corpora, tend to be much shorter than the documents in BERT's pre-train data. As a result, uniformly masking tokens at 15\% probability results in many cases where no token is being masked (e.g., about one third in LXMERT). 

Second, we note that 45\%--50\% of the masked tokens are stop-words or punctuation. While this seems a common phenomena also in text-only datasets, we show that this causes the image to be under-used in MLM pre-training for VLP. Evidently, for the popular LXMERT model, we find that the MLM validation accuracy on stop-words and punctuation is almost perfect even when omitting the image. 

To address these limitations, we propose alternative strategies aiming to mask words that require the image (e.g., physical objects). We pre-train the LXMERT model with these strategies and demonstrate their benefits in two evaluation setups.
First, on three VLP downstream tasks (GQA, \citealp{hudson2019gqa}; VQA, \citealp{goyal2017making}; NLVR2, \citealp{suhr-etal-2019-corpus}), our masking strategies consistently improve over the traditional MLM, especially in low resource settings. 
Second, we experiment with prompt based object detection~\cite{radford2021learning}, a probing task designed to elicit image objects by presenting the pre-trained models with prompts such as ``A photo of [MASK]'' and compare their top predictions with image objects. Our results show that our strategy substantially improves over the baseline sampling approach, even when trained over only a third of its epochs and half of its training data.

In our analysis, we introduce a new metric (\lossgap) to estimate the necessity of the image for a masked word during MLM. We extract the \lossgap{} value for each token in LXMERT validation pre-train data. We then present a hierarchy of semantic classes ranked by this metric, and find that the frequently masked tokens in our strategies indeed increase the image necessity.

Our main contributions are: (1) We show that the current MLM pre-training method is sub-optimal for VLP, and it does not make efficient use of pre-train data.
(2) We propose alternative masking strategies, and show that models trained with these strategies outperform the baseline strategy in two evaluation setups, especially in low resource settings.
(3) We introduce the \lossgap{} metric, which aims to explain the relation between a masked token and the image; we publicly release the computed values of this metric for the LXMERT validation set; this data may be used in future work to devise improved masking strategies.

\section{Limitations of MLM Approaches for Vision and Language}
\label{sec:current_approach}
\label{sec:current_approach}
In this section, we present the limitations of the MLM approach to vision and language tasks. We start by reviewing the way MLM is currently applied in cross-modal models, and analyzing the pre-train datasets used by most models. We observe the following two major limitations in the current approach: (1) no token is masked in roughly a third of the sentences; (2) a substantial part of the masked tokens are stop-words or punctuation, which can be predicted based on textual context alone, and do not require the image.

\subsection{Background}
Multiple studies have been proposed to modify the MLM objective in text-only domains \cite{joshi-etal-2020-spanbert, sun2019ernie, clark2020electra,Levine:2021}. However, less research has been dedicated to the implications of MLM in vision and language tasks. 

\citet{shin2021perspectives} recently reviewed how the transformer architecture \cite{vaswani2017attention} has been incorporated into vision-language cross-modal tasks. They show that most VLP models perform MLM in the same way as introduced in BERT \cite{devlin2018bert} for text-only data, randomly masking tokens with 15\% probability. Further, virtually all models are pre-trained on a handful of pre-training cross-modal datasets, including Conceptual Captions (CC; \citealp{sharma2018conceptual}); SBU captions \cite{ordonez2011im2text} and the LXMERT pre-train dataset, which is a combination of COCO \cite{lin2014microsoft}, Visual Genome \cite{krishna2017visual}, VQA \cite{goyal2017making}, VG-QA \cite{zhu2016visual7w}, and GQA \cite{hudson2019gqa}.

Importantly, all these datasets consist of $<$sentence, image$>$ pairs, where the sentence is usually a caption describing the image or, in VQA, an image-related question.

\subsection{Limitations}
\paragraph{In many cases, no token is masked.} 
Image captions tend to be shorter than the documents in BERT pre-train data, such as Wikipedia articles. BERT input sequence length is 512 tokens, while in VLP datasets the sequence length is $\approx$20 tokens.
For this reason, when masking 15\% of the tokens in the VLP models, there are cases where \textit{no token} is masked. For example, in LXMERT we find that in 36\% of the sentences, no token is masked.

\paragraph{Many masked words are stop-words and punctuation.}
\label{sec:image_nec}
We observe that over 45-50\% of tokens masked by either LXMERT, CC, and SBU are stop-words or punctuation marks.\footnote{We used nltk and gensim stop words lists.} We now describe an experiment that shows that this distribution causes the image to be under-utilized during MLM pre-training.

We follow the approach of amnesic probing \cite{elazar2021amnesic}. The intuition is that if the image is being used for cross-modal MLM, then the removal of the image should negatively influence the ability of the model to solve the task. If the removal of the image has little or no influence on the ability to solve cross-modal MLM, then the image is not a contributing factor in this task.

\begin{figure*}[!hbt]
\centering
\begin{minipage}{0.40\textwidth}
\includegraphics[width=\textwidth]{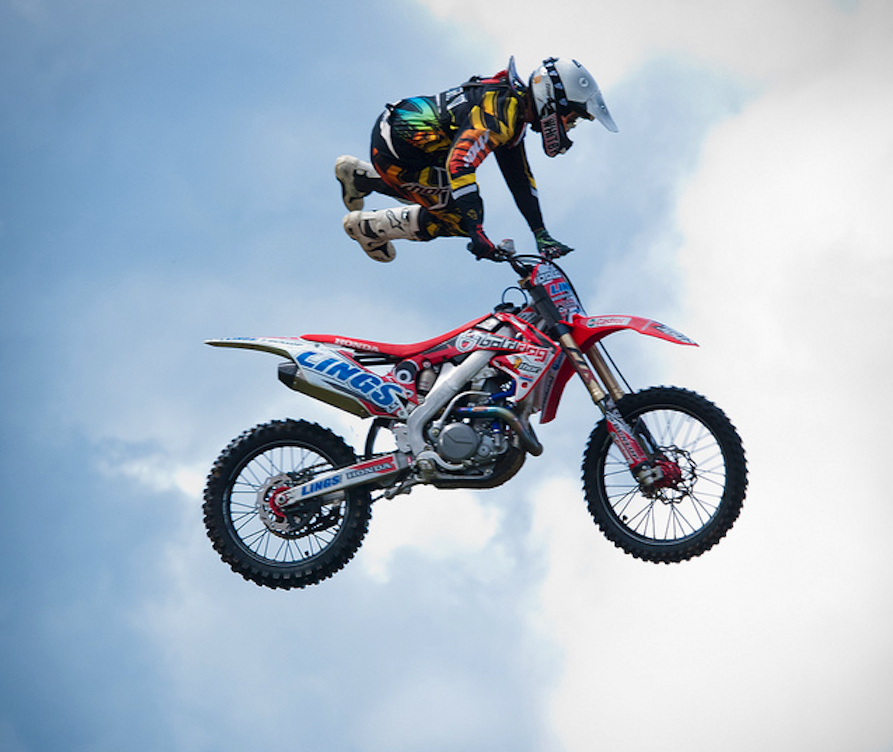}
\end{minipage}
\begin{minipage}{0.5\textwidth}
\centering
\captionsetup{type=table} 
\scalebox{0.72}{
\begin{tabular}{@{}ll@{}}
\toprule
Sentence                    & A person performs a stunt jump on a [MASK]. \\ 
Masked token                & motorcycle                                      \\
Top 5 predictions           & motorcycle, bike, ramp, bicycle, cycle          \\
Top 5 predictions w/o image & building, wall, beach, field, street            \\
Loss                        & 0.25                                        \\
Loss w/o image              & 3.96                                         \\
\lossgap{}                    & 3.71                                         \\ \bottomrule
\end{tabular}
}
\end{minipage}
\caption{An example from the extracted \lossgap{} data. The masked word is \textit{motorcycle}. Model predictions (``Top 5 predictions'') are better correlated with the image when it is given, and the loss is 0.25. Without the image, the predictions (``Top 5 predictions w/o image'') are tokens that do not appear in the image, and the loss is much higher (3.96). The \lossgap{} is the gap: 3.71.}
\label{fig:fig_motorcycle}
\end{figure*}

We consider the published pre-trained LXMERT model.\footnote{\url{https://github.com/airsplay/lxmert}} We evaluate it at inference time with the MLM task twice: with and without the image,\footnote{Without the image, we block access to the image and use the model as a single-stream model, without the co-attention layers from the image to the text. The model receives only the text and needs to complete the masked tokens.} using different masking strategies. We use the LXMERT pre-train validation data ($\approx$214K sentences). To estimate the image necessity for a masked token during MLM, we introduce the \lossgap{} metric, which is the difference in validation loss of the model prediction with and without the image. 
For example, in Figure~\ref{fig:fig_motorcycle}, the loss \emph{without the image} for predicting ``motorcycle'' is 3.96, and the loss with the image is 0.25, the \lossgap{} is 3.71.
In addition, we report the \emph{Accuracy@5} metric, which is whether the label is among the top 5 most confident predictions of the model. We compare three masking strategies, keeping a 15\% probability to mask a token: (1) \originalstrategy{} masking strategy, where a token is masked uniformly at 15\% probability; (2) masking only stop-words and punctuation; and (3) masking only content words, which is the complementary group of stop words and punctuation.

Results are presented in Table~\ref{tab:perplexity}. We observe that the model validation accuracy on stop-words and punctuation is almost perfect (96\%) even without the image. 
On the other hand, in the case of content words, accuracy is much lower without the image, and adding it increases accuracy by roughly 20\%.

\begin{table*}[!hbt]
\resizebox{\textwidth}{!}{%
\begin{tabular}{@{}ccccccc@{}}
\toprule
\textbf{Masking   strategy}         & \multicolumn{2}{c}{\textbf{With   Image}}              & \multicolumn{2}{c}{\textbf{Without   Image}}           & \multicolumn{2}{c}{\textbf{Image Necessity}}           \\ \midrule
\textbf{Metric}                     & \textbf{image loss (exp)} & \textbf{Accuracy @ 5} & \textbf{image loss (exp)} & \textbf{Accuracy @ 5} & \textbf{\lossgap{} (exp)} & \textbf{Accuracy @ 5} \\ \midrule
\textbf{\originalstrategy{}}                     & 3.2                            & 89\%                  & 8.9                            & 78\%                  & 5.7                            & 10\%                  \\
\textbf{Stop-words   \& punctuation, 15\%} & 1.5                            & 98\%                  & 2.9                            & 96\%                  & 1.4                            & 2\%                   \\
\textbf{Content words, 15\%}              & 9.4                            & 76\%                  & 38.7                           & 56\%                  & 29.3                           & 20\%                 
\\ \bottomrule
\end{tabular}
}
\caption{Performance of the LXMERT model on the MLM task, when different words are masked, with and without the image. Accuracy on stop-words and punctuation is almost perfect even when no image is present. However, for content words, the image does contribute to increased accuracy.}
\label{tab:perplexity}
\end{table*}

\section{Alternative Masking Strategies}
\label{sec:alternative_masking_strategies}
To overcome the limitations presented in the previous section, we introduce several alternative masking strategies for cross-modal MLM. The proposed strategies use several semantic classes, which are introduced in Section~\ref{sec:semantic_classes}, and then used in Section~\ref{sec:proposed_strategies}. 

\subsection{Semantic Classes}
\label{sec:semantic_classes}

\paragraph{\textit{Objects}, \textit{Attributes}, and \textit{Relationships}}
We use the definitions of \textit{objects}, \textit{attributes}, and \textit{relationships} as described in Visual Genome \cite{krishna2017visual}. \textit{Objects} represent physical entities in the image (e.g., a tiger, or a carrot). \textit{Attributes} are properties of objects, such as colors or physical state (e.g., upright). Finally,  \textit{relationships} connect between two objects. These can be actions (e.g., a tiger is \textit{eating} a carrot), spatial relations (e.g., the tiger is \emph{behind} the carrot), etc.

In order to mask the tokens that belong to those semantic classes, we first need to identify them in a given sentence.  Some datasets (e.g., GQA) include scene-graph annotations of these classes for each image. We use the annotations as ground-truth and develop heuristics to identify them automatically. For example, an \textit{Object} can be reliably annotated by identifying nouns which are also in the Visual Genome objects list. This simple heuristic achieves an accuracy of $\approx$90\% and recall of $\approx$97\% for ientifying objects on the LXMERT pre-train dataset.\resolved{\roy{accuracy/recall in identifying Objects I assume? say this explicitly}} We elaborate on these heuristics in Appendix~\ref{sec:using_obj_att_rel}.

\paragraph{Concreteness}
We hypothesize the image contributes more when predicting concrete concepts (e.g., tiger) compared to abstract concepts (e.g., hunger). 
To that end, we use a dataset of lexical concreteness presented in \cite{brysbaert2014concreteness}. This dataset provides concreteness scores (on a scale of 1-5) for over 91\% of the lemmas in LXMERT pre-training dataset.

\subsection{Proposed Strategies}
\label{sec:proposed_strategies}

We consider the following masking strategies:
\begin{itemize}
\itemsep0em 
    \item \emph{\originalstrategy{}}:
    the original masking strategy as defined in the LXMERT paper, 15\% random token masking.
    \item \emph{Objects}: Randomly mask one object word.\footnote{In $>97.2\%$ of the sentences there is at least one object. In other cases, we mask a word at random.}
    \item \emph{\cwstrategy{}}: Mask exactly one word in each sentence. Instead of almost 50--50 partition between masking stop-words and content words, increase the probability to mask content word to 80\%.
    \item \emph{Top concrete}: Mask one of the top concrete words in the sentence, weighted by their order.\footnote{Of the three words with the highest concreteness value in the sentence, mask the most concrete word with 55\% probability, the second most concrete with 30\% probability, and the third most with 15\% probability.}
    \item \emph{Stop-words \& punctuation}: as baseline, mask only stop-words \& punctuation, keeping a 15\% probability of masking. 
    \item \emph{Random 1 word}: An ablation of masking a single random word. 
\end{itemize}

\textbf{Tokenization}:
The words in the sentences are tokenized using BERT tokenizer. For strategies requiring word-level masking (Objects, Content words, Top concrete, \originalstrategy{}, Random 1 word), we mask all of the corresponding word-pieces (e.g., ``A tiger is eat \#ing'' is masked as ``A tiger is [MASK] [MASK]'').

\section{Experiments}
To evaluate the value of our proposed strategies, we conduct experiments by pre-training models with different masking strategies and evaluate them on two evaluation setups. We describe the experimental setups below.

\subsection{Downstream Tasks}
\label{sec:downstream_tasks}

\paragraph{Experimental setup} We pre-train the LXMERT architecture with the proposed masking strategies, experimenting with increasing amounts of pre-training data (10\%, 20\%, 50\%, 100\%), training for 7 epochs.\footnote{While the published LXMERT model was pre-trained for 20 epochs, we pre-train for 7 epochs because we conduct multiple pre-train experiments, and prefer to spend our budget on more experiments than a few very expensive ones.}
All other hyper-parameters are the same as the original implementation. We only modify the MLM objective, fine-tuning on three downstream tasks (VQA, GQA, NLVR2). 
For VQA and GQA, we report the mean of two experiments with different random seeds. The NLVR2 dataset is smaller ($\approx$10\% of GQA), so we report three experiments with different random seeds. Following common practice~\citep{tan-bansal-2019-lxmert}, we test GQA on the \textit{test-dev} split; NLVR2 on the public test set \textit{test-P}; and VQA on the \textit{minival} split. See corresponding papers for more details.

\begin{figure}[tb!]
\centering
\begin{minipage}{0.40\textwidth}
\includegraphics[width=\textwidth]{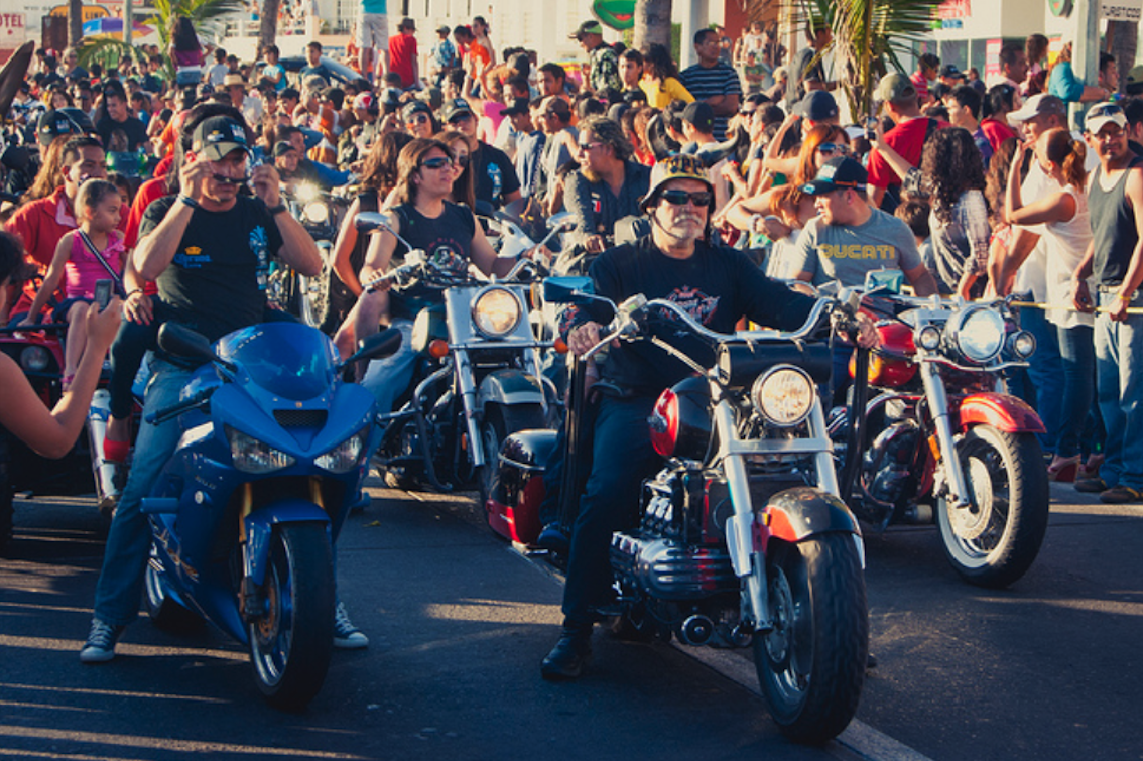}
\end{minipage}
\begin{minipage}{0.5\textwidth}
\centering
\captionsetup{type=table} 
\scalebox{0.65}{
\begin{tabular}{@{}ll@{}}
\toprule
Published LXMERT                     & \myul[red]{bathroom}, \myul[red]{beach}, \myul[red]{city}, \myul[red]{kitchen}, \myul[green]{woman} \\
Objects           & \myul[green]{motorcycle}, \myul[red]{bathroom}, \myul[green]{parade}, \myul[green]{man}, \myul[green]{crowd}          \\
 \midrule
 Ground truth objects           & glasses, gang, motorcycle, shirt, man, parade, ...          \\
 \bottomrule
\end{tabular}
}
\end{minipage}
\caption{Example of top 5 predictions for the prompt based object detection task, for the prompt ``A photo of a [MASK]''. Green underline indicate that the model predicted an object that appear in the ground truth objects (obtained from the scene graph). The model trained with \emph{Objects} masking strategy is more responsive to the image content compared to the baseline model.}
\label{fig:fig_parade_comparison}
\end{figure}

\paragraph{Results} Figure~\ref{fig:res_vqa_gqa_nlvr} presents our downstream tasks results.\footnote{Results tables presented in Appendix~\ref{sec:full_results}.} 
For brevity, we focus on the \textit{Objects} masking strategy, though the trend is similar for the other alternative strategies. We observe that our alternative masking strategies consistently outperform the \textit{\originalstrategy{}} strategy, especially in low resource settings. 
Pre-training with the \textit{Objects} strategy yields gains of 0.72--0.86\% on VQA and GQA, and 4\% on NLVR2 with 10\% of the pre-train data; 0.64--0.95\% gains on VQA and GQA, and 1.35\% on NLVR2 with 20\%; 0.5--1.02\% gains on VQA and GQA, and 1.6\% in NLVR2 with 50\%. With 100\%, the improvement is minor in GQA, VQA, but still noticeable (1.08\%) on NLVR2 (The \cwstrategy{} strategy achieves 0.49 gain on GQA with 100\%). \footnote{Preliminary experiments show that increasing the number of epochs leads to smaller gains, which emphasizes the benefits of our method in low resource settings.}

\paragraph{Ablation studies} The gains observed when using our proposed strategies can result from both changes we made to address the limitations of standard MLM presented in Section~\ref{sec:current_approach}: masking a single word in each sentence (rather than not masking any word in some cases) and deciding which word to mask (rather than randomly masking tokens). To isolate the contributing factors, we design additional experiments. We pre-train with 10\% and 20\% of the data with the \textit{random 1 word} strategy, and present the mean accuracy on the VQA and GQA in Figure~\ref{fig:ablations}. We see that this strategy outperforms the \textit{\originalstrategy{}} strategy, but under-performs \textit{Objects}. In addition, in Appendix~\ref{sec:ablation} we show experiments of varying masking probabilities rather than the baseline's 15\%, with and without multiple masked tokens per sentence, and allowing sentences without any masked token. Out of all tested settings, masking a single word achieves the best downstream results.
We conclude that the benefit of our proposed strategies comes from both choosing a single word to mask, and masking tokens that are more important. 

For completeness, we experiment with the \textit{stop-words \& punctuation} strategy with 10\% and 20\% of the data on VQA and GQA.
As expected, this strategy under-performs the \textit{\originalstrategy{}}; by 1.4\% when pre-training with 10\% of the data, and 3.37\% with 20\% the data.

\begin{figure}[hbt!]
\centering
     \includegraphics[width=\columnwidth]{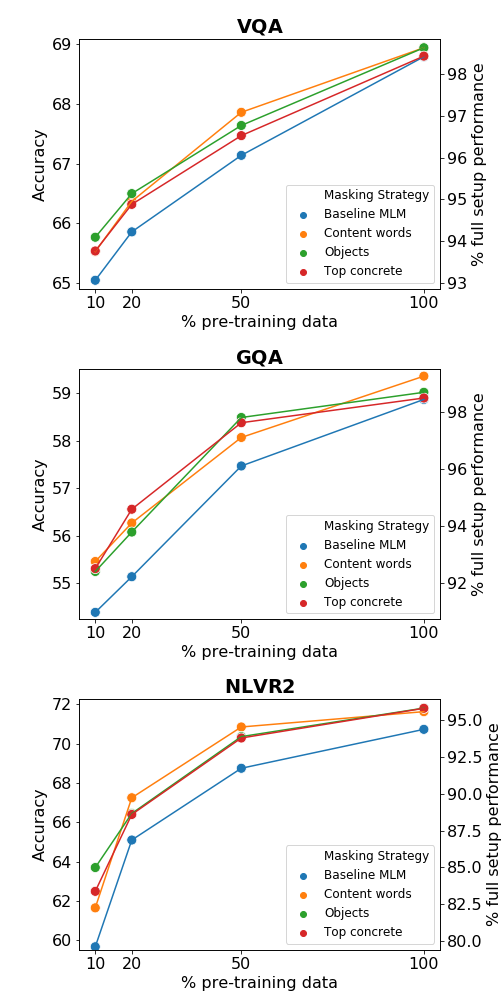}
      \caption{VQA, GQA and NLVR2 downstream tasks results for models with different masking strategies and increasing amounts of pre-train data. The left Y axis describes the accuracy, the right Y axis describes the percentage of the full setup performance (trained with 20 epochs and 100\% of the pre-train data).
      Our alternative masking strategies consistently improve over the \originalstrategy{} masking strategy, especially in low resource settings.
     \resolved{\roy{missing STDs}}
      }
      
       \label{fig:res_vqa_gqa_nlvr}
\end{figure}

\begin{figure}[hbt!]
\centering
     \includegraphics[width=0.8\columnwidth]{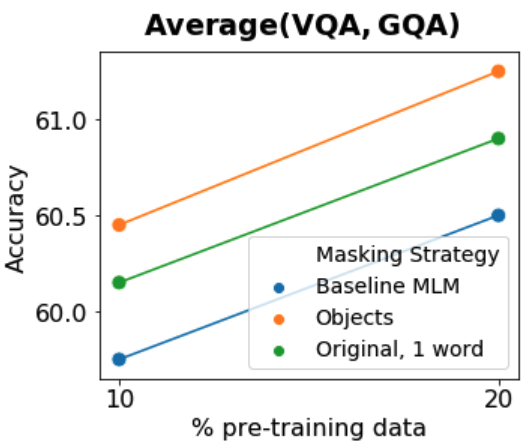}
      \caption{Ablation results for randomly masking a single word. The plot shows the average results for GQA and VQA. A model that masks a single word outperforms one with the original strategy of randomly masking 15\% of the tokens, but under-performs a model that  masks a single \textit{object} word. We conclude that the gain of our proposed strategies comes from both masking a single word, and selecting tokens that are more important.
      }
       \label{fig:ablations}
\end{figure}

\subsection{Prompt Based Object Detection}
\label{sec:prompt_base}

\begin{figure*}[h]
\centering
     \includegraphics[width=\textwidth]{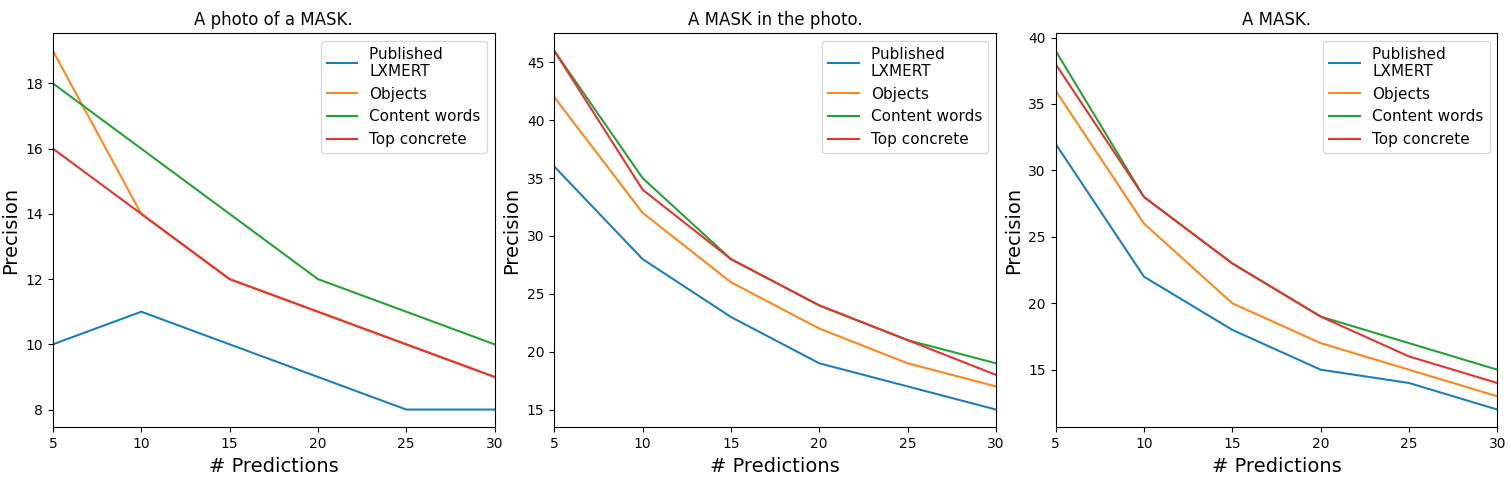}
      \caption{Precision/recall curve for prompt-base object detection task. Our models substantially improve over the published LXMERT, despite training over only a third of its epochs and half of its training data.}
       \label{fig:image_classification}
\end{figure*}

To further examine the value of our proposed masking strategies, we
examine in what way the pre-trained models trained with different strategies differ. To do so, we use prompts, and study whether a model trained for only completing \emph{Objects} (for example) will be more responsive to the image contents compared to the baseline model.

For example, given the image in Figure~\ref{fig:fig1}, we can query the model using the prompt ``A photo of a [MASK]'', and count how many of the objects  (``tiger'', ``carrot'') are in its top $k$ predictions. We compare our alternative pre-trained models, pre-trained on 50\% of the data, with the original pre-trained LXMERT model. We evaluate them on 2193 images from the LXMERT \textit{minival} split, which the model did not observe during pre-training. Given a (prompt, image) pair,  we intersect each model's top $k$ predictions with the ground-truth objects list obtained from the image ground truth scene-graph, available for these images.
We use several prompts: ``A photo of a [MASK]'' (inspired by CLIP \cite{radford2021learning}), ``A [MASK] in the photo'', and ``A [MASK]''. We present a precision for different values of $k$ in Figure~\ref{fig:image_classification}. 

Our models achieve improved precision score over published LXMERT, despite training over only a third of its epochs and half of its training data. The precision metric is simply the number of correct predictions (intersection of predictions with ground-truth objects), divided by the number of predictions. For example, when considering five top predictions ($k$=5), the published LXMERT achieves 10\% precision, compared to 18\% precision for the model trained with \emph{Content words} masking strategy. When $k$=10, the improvement is 11\% $\rightarrow$ 16\%,\resolved{\roy{do you mean 11 -> 16? you have a backslash that I think shouldn't be there}} etc. Additional results and ROC curve are available in Section~\ref{sec:full_results} in the Appendix. Our results indicate that our proposed models are more responsive to the image compared to the model trained with the \originalstrategy{} strategy. An example comparing the \originalstrategy{} model and model trained with \emph{Objects} masking strategy is presented in Figure~\ref{fig:fig_parade_comparison}. Four of the top five predictions of the model trained with \emph{Objects} masking strategy appear in the  list of ground-truth objects, while the model trained with \originalstrategy{} strategy predicts only one of the ground-truth objects.

\section{Analysis and Discussion}
\label{sec:analysis}

\subsection{Hierarchy of Masked Semantic Classes}
\label{sec:hierarchy}
\begin{figure*}[!hbt]
\centering
\newcommand{\figlen}[0]{\columnwidth}
    \includegraphics[width=\textwidth]{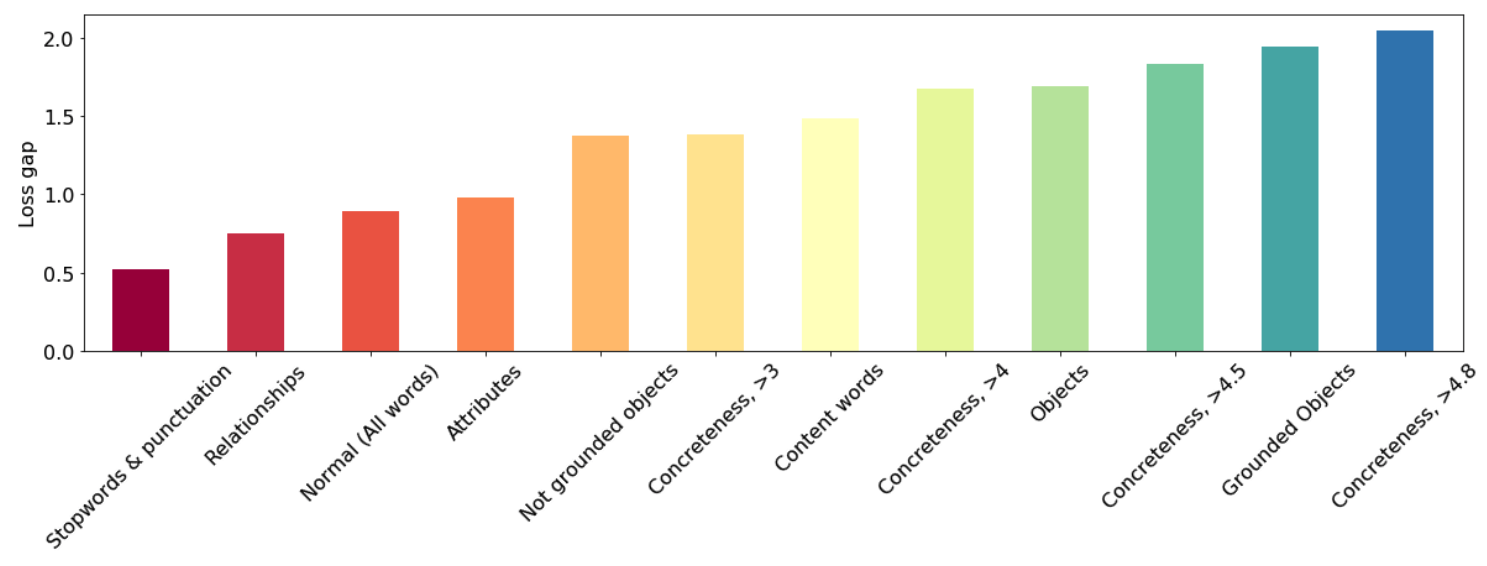}\\
  \caption{Hierarchy of semantic classes and its importance by the \lossgap{} metric (Loss without image - Loss with image).}
    \label{fig:spectral_bar}
\end{figure*}

We have shown that our strategies improve results over the \originalstrategy{}. In this section, we aim to understand if the tokens we mask make the model actively rely on the image. For this purpose, we extract the image necessity for a masked token using the \lossgap{} metric (see Section~\ref{sec:image_nec}) for every token. We use the original LXMERT pre-trained model and validation data. For each sentence, we iterate over each token, mask and predict it with and without the image. An example from the extracted \lossgap{} data is presented in Figure~\ref{fig:fig_motorcycle}.\footnote{We publish this extracted data for future work.}
Following, Figure~\ref{fig:spectral_bar}  presents a hierarchy of the different semantic classes described in Section~\ref{sec:semantic_classes}, ranked by their \lossgap{}.\footnote{The groups are not mutually exclusive.} 

We draw several observations based on that plot. First, we note that objects that appear in both text and the scene graph (dubbed grounded objects, e.g., ``tiger'') are more important than non-grounded objects. Our intuition is that grounded concepts have higher \lossgap{} compared to non-grounded concepts, as the model benefits from masking the latter. For example, consider the sentence ``Is there a \textit{tiger} in the image?'', for an image without any tiger (i.e., \textit{tiger} is not grounded). In this case, the model would not have the ability to differentiate the true word (\textit{tiger}) from any other object in the vocabulary that is also not in the image.

In addition, we observe that the objects semantic class is the most important one. We see a connection between the hierarchy and downstream performance obtained by our different strategies. \textit{Stop-words \& punctuation} are ranked the lowest, and indeed pre-training with the \textit{Stop-words \& punctuation} strategy achieves the lowest results. The strategies of \textit{Objects} and \textit{Top concrete} are ranked high, and indeed they achieve improved results compared to the \originalstrategy{}. 

\subsection{MLM Performance across Word Classes}
Many works~\cite{lu2019vilbert, tan-bansal-2019-lxmert, chen2020uniter} assume that a VLP model should include an MLM component that is capable of predicting \textit{every} masked token, including objects, properties, but also stop words and punctuation. Does a model that uses our \textit{Objects} strategy,  and masks only objects, learn to complete words from other classes? If not, can such a pre-training strategy be effective?

To examine this questions, we extend the experiment described in Section~\ref{sec:current_approach} to additional masking strategies, comparing between the different models pre-trained on 50\% of the data. Results are presented in Table~\ref{tab:val4models}. We see that the model trained with the \emph{\originalstrategy{}} masking strategy is able to complete masked words from different classes (performance are above 70\% for all cases).
\resolved{\roy{the term ``is able to mask'' is ill-defined. Talk about performance levels, as you do below for objects.}\yonatan{fixed, I ment "able to complete masked words" not "able to mask". I also added "performance are above 70\% for all cases".}} However, the model trained with \textit{Objects} masking strategy indeed learned to complete only objects. Nonetheless, its downstream performance is in fact higher than the \textit{\originalstrategy{}} model. We conclude that a model does not necessarily need to be able to complete all semantic classes, and some classes are more beneficial than others. For example, the \textit{Objects} model's performance is quite low on both completing stop-words (4\%), which is considered an easy task, and on attributes (22\%). 

A possible explanation for these findings might be that the model is evaluated mostly on retrieving objects, and had we tested it on other classes, its performance would have substantially decreased. To test this hypothesis, we inspect the same model's performance on questions with answers from different semantic types. To do so, we experiment with the GQA dataset, which includes partitioning of the answers into different semantic types, including \textit{Objects}, \textit{Relations} (subject or object of a described relation, e.g., ``what is the girl wearing?"),\resolved{\roy{I would say this is an object question (the answer is an object)}\yonatan{That's not object detection. The question requires object detection skills, but it is more extensive than simply object detection}} and  \textit{Attributes} (the properties or position of an object).

The results for the semantic type partition are presented in Table~\ref{tab:gqa_semantic_subset}. Comparing between the models trained with \textit{Objects} and \textit{\originalstrategy{}} masking strategies, the \textit{Objects} masking strategy achieves improved performance in \textit{Relationships} and \textit{Attributes}, although it never masked these kinds of tokens, and its MLM performance on these classes is considerably lower. It seems that masking only objects might assist the models to learn additional semantic classes. 

\begin{table*}[!hbt]
\centering
\begin{tabular}{@{}ccccc@{}}
\toprule
\textbf{Model}            & \multirow{2}{*}{\textbf{\begin{tabular}[c]{@{}c@{}}\originalstrategy{}\end{tabular}}} & \multirow{2}{*}{\textbf{Objects}} & \multirow{2}{*}{\textbf{\begin{tabular}[c]{@{}c@{}}\cwstrategy{}\end{tabular}}} & \multirow{2}{*}{\textbf{Top concrete}} \\ 
\textbf{Masking Strategy} &                                                                                                 &                                   &                                                                                                             &                                        \\ \midrule
\originalstrategy{}                    & 87\%                                                                                             & 27\%                               & 70\%                                                                                                         & 36\%                                    \\
Stop-words \& punctuation, 15\% & 98\%                                                                                             & 4\%                                & 80\%                                                                                                         & 13\%                                    \\
Content words, 15\%             & 74\%                                                                                             & 57\%                               & 62\%                                                                                                         & 62\%                                    \\
Objects                   & 76\%                                                                                             & 85\%                               & 82\%                                                                                                         & 83\%                                    \\
Attributes                & 70\%                                                                                             & 22\%                               & 59\%                                                                                                         & 50\%                                    \\
Relationships             & 89\%                                                                                             & 15\%                               & 75\%                                                                                                         & 25\%                                   
\\ \bottomrule
\end{tabular}
\caption{MLM Validation Accuracy@5 for different pre-training strategies, tested on different masking strategies. Interestingly, the model trained with \textit{Objects} strategy achieves low performance on all semantic classes except objects, but still achieves improved results compared to the model trained with \originalstrategy{} strategy.}
\label{tab:val4models}
\end{table*}

\begin{table}[]
\centering
\resizebox{\columnwidth}{!}{
\begin{tabular}{@{}llll@{}}
\toprule
\multirow{2}{*}{\textbf{\begin{tabular}[c]{@{}l@{}}Question   \\      semantic type\end{tabular}}} & \multirow{2}{*}{\textbf{\# Questions}} & \multicolumn{2}{l}{\textbf{Masking   Strategy}} \\ 
                                                                                                   &                                       & \textbf{\originalstrategy{}}       & \textbf{Objects}      \\ \midrule
Objects                                                                                            & 778                                   & 86.89                   & 87.79                 \\
Attributes                                                                                         & 5,186                                 & 63.17                   & 63.96                  \\
Relations                                                                                          & 5,308                                 & 49.72                   & 50.47                
\\ \bottomrule
\end{tabular}}
\caption{GQA semantic types partition performance. The model trained with \textit{Objects} masking strategy achieves improved performance compared to the baseline model on \textit{Relationships} and \textit{Attributes}, although it never masked these kind of tokens.}
\label{tab:gqa_semantic_subset}
\end{table}

\section{Related Work}
\label{sec:related_work}

\subsection{Vision Language Pre-training (VLP)} 
Recently, many VLP models have been proposed \cite{lu2019vilbert, tan-bansal-2019-lxmert, chen2020uniter}. The pre-training objectives in many cases are: (1) Masked language modeling (MLM), where a model predicts masked tokens given the sentence and the image. (2) Masked region modeling (MRM), where the model predicts masked visual object features, and (3) Sentence-image matching, where the model predicts whether the sentence belongs to the image. Some models also add the visual question answering objective during the pre-training phase~\citep{tan-bansal-2019-lxmert, li2021semvlp}. Previous works have found that the MLM objective is an important pre-training task affecting the quality of the learned representations \cite{chen2020uniter, huang2020pixel, hendricks2021decoupling}. However, the MRM objective was not always found to be important \cite{su2019vl, hendricks2021decoupling}, and the same for sentence-image prediction \cite{hendricks2021decoupling, li2019visualbert}. For this reason, we focus on the MLM objective. 

\subsection{Alternative MLM objectives in vision and language} 
Concurrently with our work, \citet{zellers2021merlot} presented an approach for pre-training over YouTube videos. They suggested a strategy of corrupting highly visual words in the masked language modeling task, observing that vanilla BERT-style often masks ungrounded words like ``umm'' or ``yeah''. 
We share the same motivation to mask highly visual words. 

\subsection{Challenges in VQA generalization}
\paragraph{Visual understanding} Language and vision tasks inherently demand deep understanding of both the text and the image. However, many works show that models can succeed on VQA \emph{datasets} using strong language priors, and by relying on superficial cues, and there are still challenges to overcome for tasks with more compositional structure~\cite{jabri2016revisiting,  zhang2016yin,goyal2017making,agarwal2020towards,bitton2021automatic,dancette2021beyond}.\resolved{\roy{Our naacl work is cited as arxiv}}
Balanced datasets such as VQA 2.0~\cite{goyal2017making} and GQA \cite{hudson2019gqa} have been presented to address these challenges. Novel models with richer visual representations \cite{zhang2021vinvl} were also presented, and some works tried to encourage the model to look at the ``correct'' image regions \cite{liu2021answer, yang2020object}. 

\paragraph{Bias} \citet{yang2021causal} and \citet{hendricks2018women} have shown that attention-based vision-language models suffer from bias that misleads the attention module to focus on spurious correlations in training data, and leads to poor generalization. Some examples are presented in Appendix~\ref{sec:examples}, Figure~\ref{fig:the_pain}. To mitigate the language priors bias, it may be beneficial to increase the focus on the image during pre-training.

\section{Conclusions}
We have shown that the current MLM pre-training method is sub-optimal for visual language pre-training, as this process tends to focus on stop words and punctuation, and in many cases does not mask any word in the sentence.
We proposed alternative masking strategies that better utilize the image during pre-training, for example, focusing on physical objects. We found improved results in two evaluation setups, especially in low resource settings. We introduced the \lossgap{} metric, which aims to explain the relation between a masked token and the image. Our analysis includes a hierarchy that describes the necessity of the image for different semantic classes. We publicly release the extracted data with this metric on the LXMERT pre-train validation data. Future work can use this information to devise new masking strategies, and progress towards VLP models that better leverage the visual aspect of the cross-modal tasks.

\section*{Acknowledgements}
We thank the reviewers for the helpful comments and feedback.
We thank Hao Tan for sharing the code and answering questions regarding LXMERT pre-training.\resolved{\roy{you also had some correspondence with the GQA first author, didn't you? or was in the previous paper?}} We also thank Leshem Choshen, Ronen Tamari, Shahaf Finder, and Nitzan Guetta Bitton for their valuable feedback. This work was supported in part by the Center for Interdisciplinary Data Science Research at the Hebrew University of Jerusalem, and research gifts from the Allen Institute for AI and Intel Corporation.

\bibliography{anthology,custom}
\bibliographystyle{acl_natbib}

\vfill\null
\clearpage
\appendix

\section{Appendix}
\label{sec:appendix}

\paragraph{Reproducibility}

The experiments have been performed with the LXMERT model~\cite{tan-bansal-2019-lxmert} with the public implementation.\footnote{https://github.com/airsplay/lxmert} The experiments were performed with NVIDIA RTX2080 GPUs. 

\begin{table}[htb]
\resizebox{\columnwidth}{!}{\begin{tabular}{@{}lllll@{}}
\toprule
\textbf{Pre-training data} & \textbf{10\%} & \textbf{20\%} & \textbf{50\%} & \textbf{100\%} \\ \midrule
\textbf{\# Epochs} & 7             & 7             & 7            & 7             \\
\textbf{Batch size}       & 64            & 64            & 100           & 256            \\
\textbf{\# GPUs}          & 1             & 1             & 3             & 4              \\
\textbf{Runtime} & 2 days        & 3 days        & 3 days        & 3 days         \\ \bottomrule
\end{tabular}}
\caption{Pre-training experiments configurations.}
\label{tab:reproducibility}
\end{table}

\begin{table}[]
\resizebox{\columnwidth}{!}{
\begin{tabular}{lllll}
\toprule
\textbf{Epoch} & \textbf{\originalstrategy{}} & \textbf{\begin{tabular}[c]{@{}l@{}}\cwstrategy{}\end{tabular}} & \textbf{Objects} & \textbf{Top Concrete} \\ \midrule
1              & 1.70            & 3.07                                                                                   & 3.23             & 3.26                  \\
2              & 1.46            & 2.11                                                                                   & 2.28             & 2.29                  \\
3              & 1.40            & 1.97                                                                                   & 2.14             & 2.15                  \\
4              & 1.36            & 1.88                                                                                   & 2.04             & 2.05                  \\
5              & 1.33            & 1.81                                                                                   & 1.96             & 1.98                  \\
6              & 1.30            & 1.75                                                                                   & 1.90             & 1.91                  \\
7              & 1.27            & 1.71                                                                                   & 1.84             & 1.86                  \\
8              & 1.25            &                                                                                        &                  &                       \\
9              & 1.27            &                                                                                        &                  &                       \\
10             & 1.23            &                                                                                        &                  &                       \\
11             & 1.21            &                                                                                        &                  &                       \\
12             & 1.19            &                                                                                        &                  &                       \\
13             & 1.17            &                                                                                        &                  &                       \\
14             & 1.16            &                                                                                        &                  &                       \\
15             & 1.14            &                                                                                        &                  &                       \\
16             & 1.12            &                                                                                        &                  &                       \\
17             & 1.11            &                                                                                        &                  &                       \\
18             & 1.09            &                                                                                        &                  &                      
\\ \bottomrule
\end{tabular}}
\caption{Training loss for models trained in different masking strategies. The training loss for the original is obtained from the original model repository. Because we focus on tokens that are more difficult for the model to complete, the training loss is higher.}
\label{tab:training_loss}
\end{table}

\subsection{Detection of Objects, Attributes and Relationships}
\label{sec:using_obj_att_rel}

\paragraph{Using the annotated scene-graph as ground truth}
A simple way to detect \textit{objects}, \textit{attributes}, and \textit{relationships} in captions, is to obtain it, given that the image has scene-graph annotation from Visual-Genome or GQA. In LXMERT pre-training data,  83\% of the sentences have scene-graph annotations for their corresponding image. For example, given the sentence, image pair: ``The rabbit is eating the orange carrot'', and an image, the ground truth by the scene-graph will include  \textit{Objects}: rabbit, carrot; \textit{Attributes}: orange; and \textit{Relationships}: eating. 
When obtained from the scene-graph, we call it ``Grounded'' (Grounded objects, grounded attributes, and grounded relationships). 

\paragraph{Predicting \textit{objects}, \textit{attributes}, and \textit{relationships} in each caption:} For more general and scalable method when scene-graph is not available, we can use matching heuristics. We use the Part-of-speech tagging (POS), and we aggregate lists of Objects, Attribute and Relationships from Visual Genome dataset annotations.\footnote{\url{http://visualgenome.org/api/v0/api_home.html}} Those are our heuristics:\footnote{Our full code, including code to detect the semantic type tokens will be published} 
\begin{itemize}
    \item \textit{Objects} are words with POS = ``NOUN'' and in Visual Genome objects list. 
    \item \textit{Attributes} are words with POS = ``ADJ'' and in Visual Genome attributes list. 
    \item \textit{Relationships} are words with POS = ``ADP'' or ``VERB'', and in Visual Genome relationships list. 
\end{itemize}

Those simple rules are our predictions for detecting \textit{Objects}, \textit{Attributes}, and \textit{Relationships} in a sentence. 

\paragraph{Validation of the \textit{objects}  \textit{attributes} and  \textit{relationships} task: } We can now evaluate the predicted \textit{objects}, \textit{attributes} and  \textit{relationships} with the ground-truth obtained from the scene-graph. The grounding method (matching between the caption and the scene-graph) we use is simple: exact match between the word in the scene-graph and the caption. Using a more complex grounding algorithm will not change our predictions, but it can only improve our results (For example, if the caption has ``women'' that was predicted as \textit{Object}, and the scene-graph has ``woman'', it is currently counted as ``False-Positive'' because it's not exact match). Results are presented at Table~\ref{tab:detection_obj_att_rel}. 

\begin{table}[]
\begin{tabular}{@{}cccc@{}}
\toprule
                       & \textbf{\# items} & \textbf{Accuracy} & \textbf{Recall} \\ \midrule
\textbf{Objects}       & 7,484,940         & 89.89             & 97.39           \\
\textbf{Attributes}    & 3,240,096         & 92.91             & 79.91           \\
\textbf{Relationships} & 3,195,345         & 86.42             & 96.88           \\ \bottomrule
\end{tabular}
\caption{Detection performance of \textit{Objects}, \textit{Attributes}, and \textit{Relationships}.}
\label{tab:detection_obj_att_rel}
\end{table}

\subsection{Concrete and Abstract definitions}
\label{sec:concrete_and_abstract}
The concreteness annotation dataset \cite{brysbaert2014concreteness} is annotated by 20-30 annotators. The rating scale is 1-5, when 1 is abstract, and 5 is concrete. This is how they define concrete: ``A concrete word comes with a higher rating and refers to
something that exists in reality ; you can have immediate experience of it through your senses (smelling, tasting, touching, hearing, seeing) and the actions you do. The easiest way to explain a word is by pointing to it or by demonstrating it.''

This is how they define abstract: ``An abstract word comes with a lower rating and refers to something you cannot experience directly through your senses
or actions. Its meaning depends on language. The easiest way to explain it is by using other words''. 

\section{Additional Experiments}
\label{sec:ablation}
\subsection{How good is current pre-training?}

We want to asses contribution of the current LXMERT pre-training. We conduct fine-tune experiments with LXMERT without pre-tain.
Results are presented at Table~\ref{tab:how_good_is_curr_pt}. We see that pre-training adds $\approx$6.5 in GQA, $\approx$4.8 in VQA, and $\approx$23.8 in NLVR2. 

\begin{table}[hbt!]
\resizebox{0.45\textwidth}{!}{%
\begin{tabular}{@{}|c|c|c|c|@{}}
\toprule
\textbf{Dataset}                                                                       & \textbf{GQA} & \textbf{VQA} & \textbf{NLVR2} \\ \midrule
No pre-train                                                                            & 53.24        & 65.10        & 51.07          \\ \midrule
\begin{tabular}[c]{@{}c@{}}Pre-training all data\\ Reported LXMERT GitHub results\end{tabular} & 59.80        & 69.90        & 74.95          \\ \bottomrule
\end{tabular}
}
\caption{Downstream task performance for limited pre-training methods.}
\label{tab:how_good_is_curr_pt}
\end{table}

\subsection{How to change the 15\% masking amount?}
\label{sec:how_to_change_15}

In Section~\ref{sec:current_approach} we discussed that 15\% with short captions ($\approx$6.86) causes that with third of the cases no token is masked, in another third 1 token is masked, and in the last third, multiple tokens are masked. 

We isolate those factors by conducting 3 experiments:
\begin{itemize}
    \item Not allowing 0 masked (if 0 tokens were masked, sampling 1 token to mask).
    \item Not allowing multiple masked (if multiple tokens were masked, sample 1 token from them to mask)
    \item Masking only 1 word. 
\end{itemize}

Results are presented at Table~\ref{tab:how_to_change_15}. 

\begin{table}[]
\centering
\resizebox{0.5\textwidth}{!}{%
\begin{tabular}{@{}|c|c|c|c|@{}}
\toprule
                                       & \textbf{GQA} & \textbf{VQA} & \textbf{NLVR2} \\ \midrule
\textbf{\originalstrategy{}}                        & 54.4         & 65.06        & 58.55          \\ \midrule
\textbf{Don't allow 0   masked}        & 54.98        & 65.4         & 59.45          \\ \midrule
\textbf{Don't allow   multiple masked} & 54.46        & 65           & 58.82          \\ \midrule
\textbf{Mask 1 word}                   & 55.07        & 65.26        & 61.25          \\ \bottomrule
\end{tabular}
}
\caption{Changing 15\% masking amount. Masking 1 word achieves the higher downstream tasks results.}
\label{tab:how_to_change_15}
\end{table}

We can see that not allowing multiple masked tokens helps a bit. Not allowing 0 masked tokens helps more. And masking 1 word is the better overall strategy. 

\subsection{Full results tables}
\label{sec:full_results}
\begin{table*}[!htbp]
\centering
\begin{tabular}{lllll}
\toprule
                                            & \multicolumn{4}{l}{\textbf{\% of pre-train data}}          \\
\textbf{Masking Strategy}                   & \textbf{10} & \textbf{20} & \textbf{50} & \textbf{100} \\ \midrule
\textbf{\originalstrategy{}}                                                                      & 65.05 $_{\pm0.02}$ & 65.86 $_{\pm0.06}$ & 67.14 $_{\pm0.2}$  & 68.79 $_{\pm0.02}$ \\
\textbf{\begin{tabular}[c]{@{}l@{}}\cwstrategy{} \end{tabular}} & 65.53 $_{\pm0.04}$ & 66.37 $_{\pm0.04}$ & 67.86 $_{\pm0.08}$ & 68.94 $_{\pm0.05}$ \\
\textbf{Objects}                                                                       & 65.77 $_{\pm0.05}$ & 66.5 $_{\pm0.04}$ & 67.64 $_{\pm0.08}$ & 68.94 $_{\pm0.06}$ \\
\textbf{Top concrete}                                                                  & 65.54 $_{\pm0.21}$ & 66.32 $_{\pm0.02}$ & 67.47 $_{\pm0.1}$ & 68.8 $_{\pm0.03}$
\\ \bottomrule
\end{tabular}
\caption{Full VQA 2.0 results, mean$_{\pm std}$}
\label{tab:full_vqa_results}
\end{table*}

\begin{table*}[!htbp]
\centering
\begin{tabular}{lllll}
\toprule
                                            & \multicolumn{4}{l}{\textbf{\% of pre-train data}}          \\
\textbf{Masking Strategy}                   & \textbf{10} & \textbf{20} & \textbf{50} & \textbf{100} \\ \midrule
\textbf{\originalstrategy{}}                                                                      & 54.39 $_{\pm0.01}$        & 55.14 $_{\pm0.02}$        & 57.47 $_{\pm0.13}$        & 58.87 $_{\pm0.04}$         \\
\textbf{\begin{tabular}[c]{@{}l@{}}\cwstrategy{}\end{tabular}} & 55.46 $_{\pm0.04}$       & 56.27 $_{\pm0.33}$          & 58.07 $_{\pm0.09}$          & 59.36 $_{\pm0.08}$         \\
\textbf{Objects}                                                                       & 55.25 $_{\pm0.21}$        & 56.08 $_{\pm0.10}$          & 58.49 $_{\pm0.01}$        & 59.02 $_{\pm0.03}$           \\
\textbf{Top Concrete}                                                                  & 55.31 $_{\pm0.12}$       & 56.56 $_{\pm0.35}$       & 58.38 $_{\pm0.25}$        & 58.9 $_{\pm0.04}$        
\\ \bottomrule
\end{tabular}
\caption{Full GQA results, mean$_{\pm std}$}
\label{tab:full_gqa_results}
\end{table*}

\begin{table*}[!htbp]
\centering
\begin{tabular}{lllll}
\toprule
                                            & \multicolumn{4}{l}{\textbf{\% of pre-train data}}          \\
\textbf{Masking Strategy}                   & \textbf{10} & \textbf{20} & \textbf{50} & \textbf{100} \\ \midrule
\textbf{\originalstrategy{}}                                                                     & 59.67 $_{\pm1.04}$ & 65.1 $_{\pm1.13}$ & 68.75 $_{\pm0.53}$ & 70.73 $_{\pm0.65}$ \\
\textbf{\begin{tabular}[c]{@{}l@{}}\cwstrategy{}\end{tabular}} & 61.65 $_{\pm0.95}$ & 67.25 $_{\pm0.48}$ & 70.85 $_{\pm0.06}$ & 71.63 $_{\pm0.44}$ \\
\textbf{Objects}                                                                      & 63.7 $_{\pm0.14}$ & 66.45 $_{\pm1.2}$ & 70.36 $_{\pm0.91}$ & 71.81 $_{\pm0.51}$ \\
\textbf{Top Concrete}                                                                 & 62.49 $_{\pm0.72}$ & 66.4 $_{\pm0.56}$ & 70.29 $_{\pm0.22}$ & 71.8 $_{\pm0.1}$
\\ \bottomrule
\end{tabular}
\caption{Full NLVR2 results, mean mean$_{\pm std}$}
\label{tab:full_nlvr2_results}
\end{table*}

\begin{figure*}[h]
\centering
     \includegraphics[width=\textwidth]{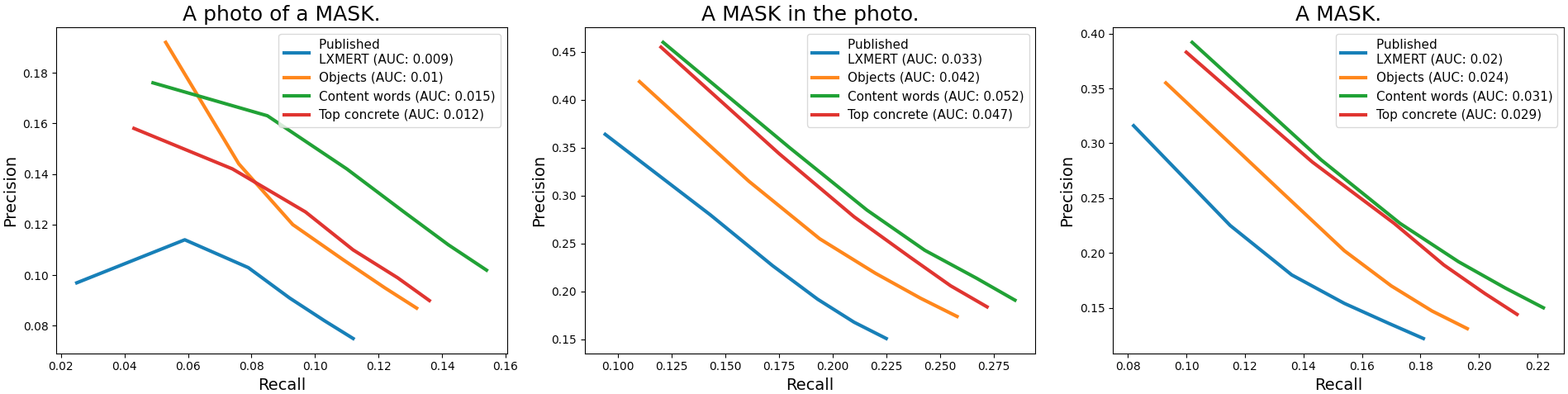}
      \caption{Precision-recall curve for prompt-base object detection task. Our models achieve improved results over published LXMERT, although trained with a half of the pre-train data and a third of the epochs.}
       \label{fig:image_classification_pr}
\end{figure*}

\subsection{Examples}
\label{sec:examples}

\begin{figure*}[]
\centering
\newcommand{\figlen}[0]{\columnwidth}
    \includegraphics[width=0.75\textwidth]{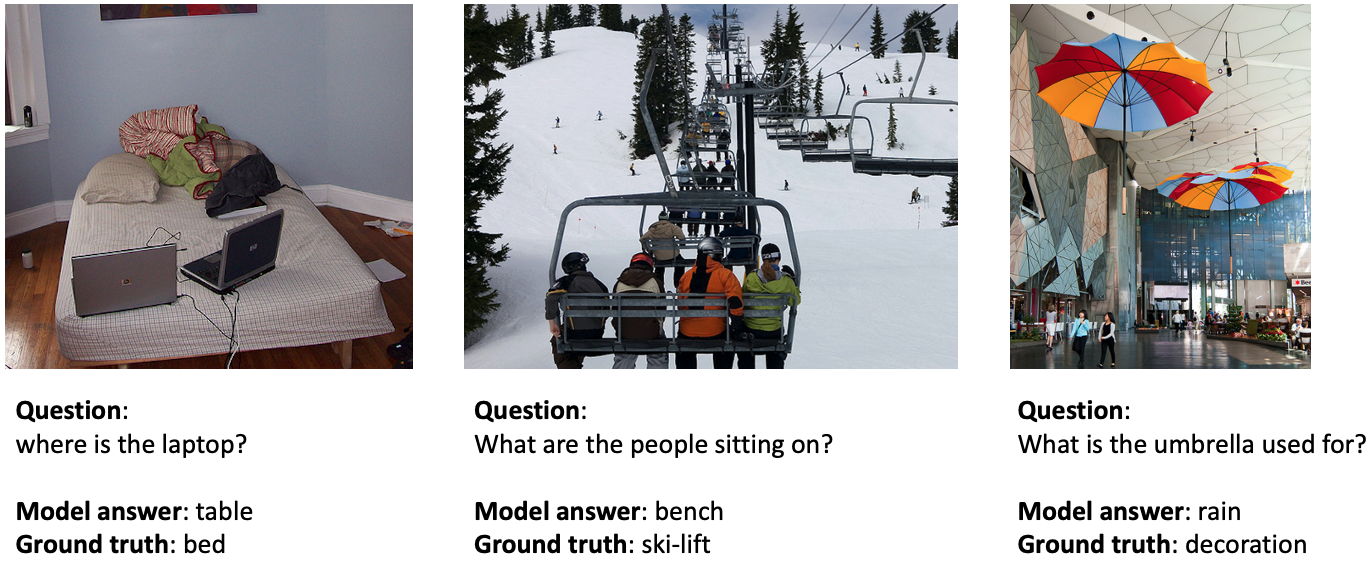}\\
  \caption{
  LXMERT mistakes observed on examples from GQA and VQA. 
  The tendency of VLP models is to predict something that is correlated with the text, or common answers. 
  In many cases, the prediction is not an item that even appears in the image.
  }
    \label{fig:the_pain}
\end{figure*}

\begin{figure*}[tb!]
\centering
\begin{minipage}{0.40\textwidth}
\includegraphics[width=\textwidth]{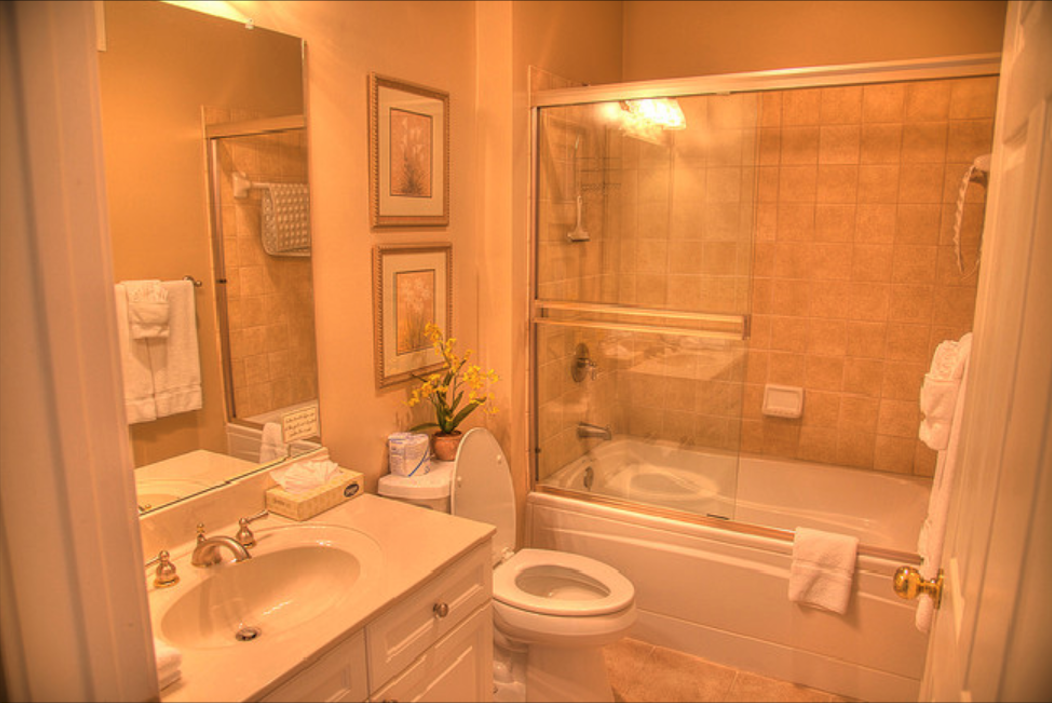}
\end{minipage}
\begin{minipage}{0.5\textwidth}
\centering
\captionsetup{type=table} 
\scalebox{0.65}{
\begin{tabular}{@{}ll@{}}
\toprule
Published LXMERT                     & \myul[green]{bathroom}, \myul[red]{kitchen}, \myul[red]{bedroom}, \myul[red]{beach}, \myul[red]{city} \\
Objects           & \myul[green]{bathroom}, \myul[red]{restroom}, \myul[green]{sink}, \myul[green]{toilet}, \myul[green]{mirror}          \\
 \midrule
 Ground truth objects           & tile, toilet, wash cloth, tub, sink, mirror, ...          \\
 \bottomrule
\end{tabular}
}
\end{minipage}
\begin{minipage}{0.40\textwidth}
\includegraphics[width=\textwidth]{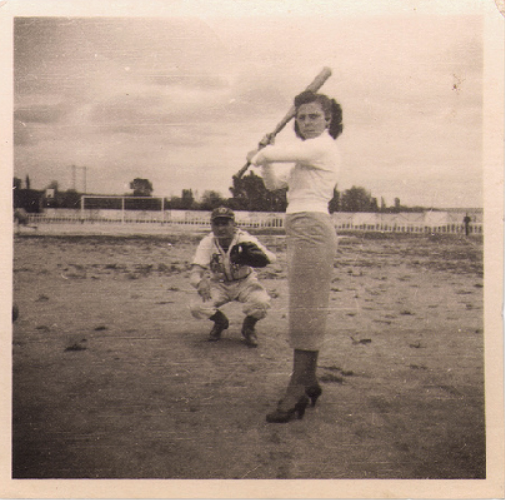}
\end{minipage}
\begin{minipage}{0.5\textwidth}
\centering
\captionsetup{type=table} 
\scalebox{0.65}{
\begin{tabular}{@{}ll@{}}
\toprule
Published LXMERT                     & \myul[red]{beach}, \myul[green]{field}, \myul[red]{bathroom}, \myul[green]{woman}, \myul[green]{man} \\
Objects           & \myul[red]{beach}, \myul[green]{field}, \myul[green]{baseball}, \myul[green]{woman}, \myul[red]{game}          \\
 \midrule
 Ground truth objects           & bat, shirt, catcher, glove, lot, distance, ...          \\
 \bottomrule
\end{tabular}
}
\end{minipage}

\caption{Additional examples of top 5 predictions for the prompt based object detection task, for the prompt ``A photo of a [MASK]''. Green underline indicate that the model predicted an object that appear in the ground truth objects (obtained from the scene graph).}
\label{fig:fig_bathroom_comparison}
\end{figure*}

\end{document}